\RequirePackage{fix-cm}

\documentclass[twocolumn]{svjour3}

\usepackage{amsmath,amsfonts}
\usepackage{algorithmic}
\usepackage{array}
\usepackage[caption=false,font=normalsize,labelfont=sf,textfont=sf]{subfig}
\usepackage{textcomp}
\usepackage{stfloats}
\usepackage{url}
\usepackage{verbatim}
\usepackage{graphicx}
\usepackage{wrapfig}
\usepackage[T1]{fontenc}
\usepackage{blindtext}
\usepackage[table]{xcolor}
\usepackage{booktabs}
\usepackage{amssymb}
\usepackage{placeins}
\usepackage{diagbox} 
\usepackage{gensymb}
\usepackage[pdfencoding=auto,
            colorlinks,
            linkcolor = blue,
            urlcolor  = blue,
            citecolor = blue]{hyperref}
\usepackage{multirow}
\usepackage{microtype}
\usepackage{enumitem}
\usepackage{flushend}

\usepackage[round]{natbib}

\let\oldcite\cite
\renewcommand*\cite[1]{(\oldcite{#1})}

\DeclareMathOperator*{\argmax}{arg\,max}
\DeclareMathOperator*{\argmin}{arg\,min}

\usepackage{mathtools}

\usepackage[colorinlistoftodos]{todonotes}

\usepackage[free-standing-units=true]{siunitx}

\newcommand{\eg}{\mbox{e.\,g.}\xspace}

\newcommand{\etal}{\emph{et~al.}\xspace}
   
\renewcommand{\[}{\begin{equation}}
\renewcommand{\]}{\end{equation}}

\renewcommand{\eqref}[1]{Eq.~(\ref{#1})}
\newcommand{\figref}[1]{Fig.~\ref{#1}}

\DeclareMathAlphabet{\mathcal}{OMS}{cmsy}{m}{n}

\renewcommand{\vec}[1]{\ensuremath{\mathbf{#1}}}
\newcommand{\mat}[1]{\ensuremath{\mathbf{#1}}}
\newcommand{\norm}[1]{\left\lVert#1\right\rVert}
\newcommand{\fd}[1]{\ensuremath{\dot{#1}}}

\newcommand{\calA}{{\cal A}}

\newcommand{\calF}{{\cal F}}

\newcommand{\calK}{{\cal K}}
\newcommand{\calL}{{\cal L}}

\newcommand{\calP}{{\cal P}}

\newcommand{\calX}{{\cal X}}
\newcommand{\calZ}{{\cal Z}}

\newcommand{\World}{\mathtt{W}}
\newcommand{\EndEff}{\mathtt{E}}

\newcommand{\Articulation}{\mathtt{A}}

\newcommand{\Identity}{\mathbf{I}}

\newcommand{\Real}{\mathbb{R}}

\newcommand{\SE}{\mathrm{SE}}
\newcommand{\SEthree}{\SE(3)}

\newcommand{\Cov}{\mathbf{\Sigma}}

\newcommand{\State}{\boldsymbol{x}}

\newcommand{\linvel}{\mathbf{v}}
\newcommand{\rotvel}{\boldsymbol{\omega}}

\newcommand{\xihat}{\hat{\xi}}
\newcommand{\part}{\mathbf{T}}

\newcommand{\partA}{\mathbf{T_{\mathtt{A}}}}
\newcommand{\partB}{\mathbf{T_{\mathtt{B}}}}
\newcommand{\partWA}{\mathbf{T_{\mathtt{WA}}}}
\newcommand{\partWB}{\mathbf{T_{\mathtt{WB}}}}

\newcommand{\partBA}{\mathbf{T}_{\mathtt{BA}}}

\newcommand{\flow}{\mathbf{f}}

\journalname{Autonomous Robots}

\begin{document}

\title{
Online Estimation and Manipulation of Articulated Objects
}

\titlerunning{Online Estimation and Manipulation of Articulated Objects}

\author{Russell Buchanan \and Adrian Röfer \and Jo\~{a}o Moura \and Abhinav Valada \and Sethu Vijayakumar}

\institute{R. Buchanan \at RIPL-Lab, University of Waterloo, Canada \\
        \email{russell.buchanan@uwaterloo.ca} \and
           A. Röfer \and A. Valada \at Robot Learning Lab, University of Freiburg, Germany \\
        \email{\{aroefer,valada\}@cs.uni-freiburg.de}
           \and
           J. Moura \and S. Vijayakumar \at School of Informatics, University of Edinburgh, UK\\
           \email{\{joao.moura,sethu.vijayakumar\}@ed.ac.uk}
}

\date{Received: date / Accepted: date}

\maketitle
\begin{abstract}
    From refrigerators to kitchen drawers, humans interact with articulated objects effortlessly every day while completing household chores. For automating these tasks, service robots must be capable of manipulating arbitrary articulated objects. Recent deep learning methods have been shown to predict valuable priors on the affordance of articulated objects from vision. In contrast, many other works estimate object articulations by observing the articulation motion, but this requires the robot to already be capable of manipulating the object. In this article, we propose a novel approach combining these methods by using a factor graph for online estimation of articulation which fuses learned visual priors and proprioceptive sensing during interaction into an analytical model of articulation based on Screw Theory. With our method, a robotic system makes an initial prediction of articulation from vision before touching the object, and then quickly updates the estimate from kinematic and force sensing during manipulation. We evaluate our method extensively in both simulations and real-world robotic manipulation experiments. We demonstrate several closed-loop estimation and manipulation experiments in which the robot was capable of opening previously unseen drawers. In real hardware experiments, the robot achieved a 75\% success rate for autonomous opening of unknown articulated objects.
\end{abstract}

\section{Introduction}
If service robots are to assist humans in performing common tasks such as cooking and cleaning, they must be capable of interacting with and manipulating common articulated objects such as dishwashers, doors, and drawers. To manipulate these objects, a robot would need an understanding of the articulation, either as an analytical model (\eg revolute, prismatic, or screw joint) or as a model implicitly learned through a neural network. Many recent works have shown how deep neural networks can predict articulated object affordance using point cloud measurements. To achieve this, common household articulated objects are rendered in simulation with randomized states as training examples. Because most common household objects have reliably repeatable articulations, such as refrigerator doors, the learned models effectively generalize to real data. However, predicting articulation from visual data alone can often be unreliable.

\begin{figure}[t]
	\centering
	\includegraphics[width=\columnwidth]{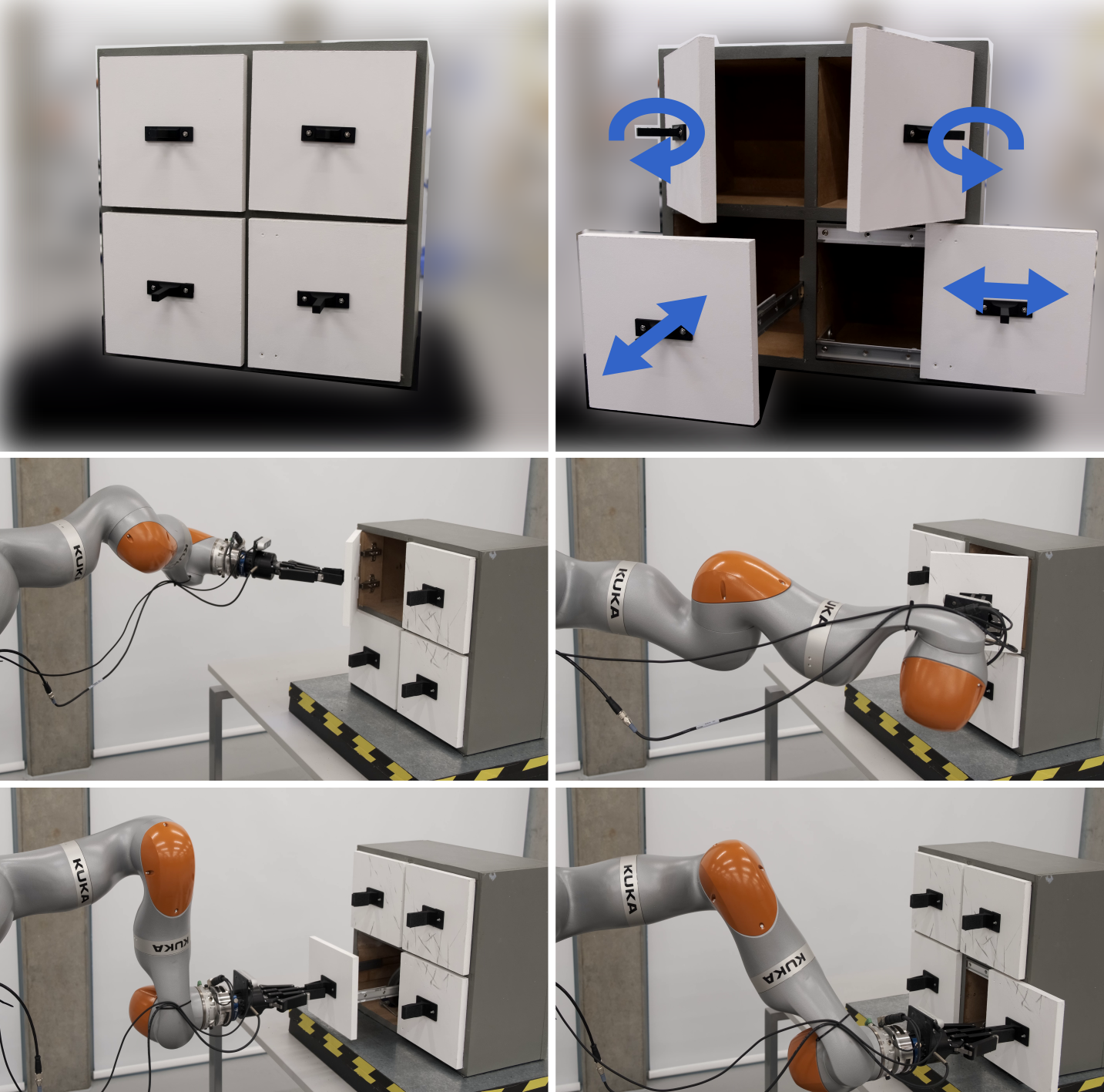}
	\caption{\textbf{Top row:} a cabinet with a set of visually identical doors. Their different articulations are only revealed once open. It would not be possible from visual inspection alone to predict how each door opens. \textbf{Middle and Bottom rows:} the robot autonomously opens each of the cabinet doors while estimating articulation online.}
	\label{fig:motivation-cabinet}
\end{figure}

For example, the cabinet in \figref{fig:motivation-cabinet} has four doors which appear identical when closed. It is impossible for humans or robots to reliably predict the articulation from vision alone. However, once a person or robot interacts with them, they are revealed to open in completely different ways. This is challenging for robotic systems that rely exclusively on vision for understanding articulations and are not capable of updating their articulation estimate online. In this work, we propose a novel method for jointly optimizing visual, force, and kinematic sensing for online estimation of articulated objects.

There is another branch of research that has focused on probabilistic estimation of articulations. These works have typically used analytical models of articulation and estimate the object articulation through observations of the motion of the object during interaction. However, these works largely rely on a good initial guess of articulation so that the robot can begin moving the object. 

In this work, we significantly improve upon our previous work presented in~\citet{Buchanan2024} in which we first investigated estimation of articulated objects. The first of these improvements is a neural network for affordance prediction which incorporates uncertainty in predictions and a completely new method of including learned articulation affordances into a factor graph to provide a good initial guess of articulation. We have also incorporated kinematic and force sensing in the factor graph which updates the estimate online during interaction. The result is a robust multi-modal articulation estimation framework. The contributions of this paper are as follows:
\begin{itemize}[topsep=0pt]
    \item We propose online estimation of articulation parameters using vision and proprioceptive sensing in a factor graph framework. This improves upon our previous work with a new uncertainty-aware articulation factor leading to improved robustness in articulation prediction.
    \item We additionally introduce a new force sensing factor for articulation estimation. 
    \item We demonstrate full system integration with shared autonomy for unseen opening articulated objects.
    \item We validate our system with extensive real-world experimentation, opening visually ambiguous articulations with the estimation running in a closed loop. We demonstrate improvements over~\citet{Buchanan2024} by opening all doors for the cabinet in \figref{fig:motivation-cabinet}, which was not previously possible.
\end{itemize}

\section{Related Work}
In this section, we provide a summary of related works on the estimation of articulated objects. This is a challenging problem that has been investigated in many different ways in computer vision, and robotics. In this work, we are concerned with robotic manipulation of articulated objects. Therefore, in the following section, we first cover related works in interactive perception~\cite{bohg_interactive_2017}, which has a long history of use for estimation and manipulation of articulated objects. Then, we briefly cover the most relevant, recent deep learning methods for vision-based articulation prediction. Finally, we discuss some methods that have integrated different systems together for robot experiments.

\subsection{Interactive Perception}
\label{sec:interactive-perception}
Interactive perception is the principle that robot perception can be significantly facilitated when the robot interacts with its environment to collect information. This has been applied extensively in the estimation of articulated objects, as once the robot has grasped an articulated object and started to move it, there are many sources of information from which to infer the articulation parameters. Today, few works use proprioception for estimating articulation due to challenges in identifying an initial grasp point and pulling direction. In 2010, \citet{Jain2010} simplified the problem by assuming a prior known grasp pose and initial opening force vector. This allowed their method to autonomously open several everyday objects, such as cabinets and drawers, while only using force and kinematic sensing. They demonstrated that once a robot is physically interacting with an articulated object and is given a good initial motion direction, proprioception alone can be sufficient to manipulate most articulated objects.

More commonly in recent research, proprioception is either fused with vision, or vision alone is used. \citet{Sturm2011} introduced a probabilistic framework for maximizing the probability of a joint type and joint parameters given an observed pose trajectory of the moving part of an articulated object (e.g., a cabinet door). To track the trajectory of the moving part, they demonstrate several different sensing methods, including visual tracking of fiducial markers, depth image-based markerless tracking, and kinematic sensing. They integrated their method with \citet{Jain2010} for real robot experiments, specifying the initial grasp point and direction of motion. 

Later work would focus on visual perception, using bundle adjustment to track visual features throughout the course of an interaction~\cite{katz_interactive_2014}. \citet{MartinMartin2022} also introduced a framework that can estimate online from vision and tactile sensing. As in previous work, they track the motion of visual features while the robot is interacting with the object. This is fused with force/torque sensing, haptic sensing from a soft robotic hand, and end-effector pose measurements. \citet{Heppert2022} proposed a visual neural network for tracking the motion of the object parts from vision. The tracked poses are connected by a factor graph to estimate the joint parameters. In their experiments, they estimated an unknown articulation; however, their controller used a prescribed motion to open the object, giving sufficient information to the estimator. 

All of these interactive perception methods require a prior grasp point and a good initial guess of the articulation. In our previous work~\cite{Buchanan2024} we used a factor graph to merge learned visual predictions with kinematic sensing. This allowed our method to automatically make an initial guess of opening direction and then update the estimate online during interaction. However, in this early work, we could only demonstrate opening of objects that require pulling motions to be opened and could not demonstrate opening of sliding doors, such as the bottom right door in Fig.~\ref{fig:motivation-cabinet}. 

In this work, like \citet{Heppert2022} and \citet{Buchanan2024},  we use a factor graph to fuse measurements of part poses to estimate a joint screw model. However, unlike these previous works, we use an uncertainty-based deep neural network prediction from visual sensing to give the robot an initial estimate of the articulation. This is then updated using both force sensing and kinematic sensing to enable the opening of any articulation, including sliding doors. Our use of both interactive perception and learning-based predictions allows us to perform closed-loop control and estimation while opening unknown articulated objects.

\subsection{Learning-Based Articulation Prediction}
Many recent works have investigated using only visual information with deep learning to predict articulation without the need for object interaction~\cite{Li2019, Jain2021, Jiang2022ECCV}. Often, these works use simulated datasets such as the PartNet-Mobility dataset~\cite{Mo2018}, which contains examples of common articulated objects. Since many common household objects have predictable articulations (e.g., refrigerator doors), these works assume that articulation can be predicted in most cases through visual inspection.

Earlier learning-based works explicitly classified objects for the prediction of articulation, but more recently, there has been a focus on learning category-free articulation affordances~\cite{mo_where2act_2021, xu_umpnet_2022, Eisner2022, Eisner2023}, which describe how a user can interact with an object without classifying the object from vision. This is typically parameterized as a normalized vector that describes the motion of a point on the articulated part of an object. \citet{Bahl2023} used a neural network to also predict grasp pose on the object as well as the opening trajectory from human demonstrations. Some recent learning-based work has incorporated interaction. \citet{Jiang2022CVPR} used point cloud data collected before and after human interaction with the target articulated object. \citet{Nie2023} introduced a method that predicts articulation as well as proposes an interaction through which to observe the motion and update the articulation estimate.

All of these learning-based works have similar limitations that hinder their use on real robots. They use only visual information and have large computational requirements, which prevents online estimation. Therefore, when they are used with real data, they typically take a single ``snapshot'' of the object and make a single inference. If there is any error in the prediction, there is no way to update the estimate. Additionally, due to the reliance on recognizing visual similarity in objects compared to past training experiences, these methods exhibit poor performance on an object like in \figref{fig:motivation-cabinet}, which has no visual indicators as to how it opens. If the predictions are wrong, then these methods are reliant on highly compliant controllers to account for the error due to a lack of online estimation. 

\subsection{Systems}
In our work, we provide not only an estimation method but also a full system for opening articulated objects with shared autonomy. Therefore, we also discuss some related work that developed systems for the manipulation of articulated objects. \citet{Mittal2021} introduced a system for whole-body mobile manipulation. They used the category-level object pose prediction network from \citet{Li2019}. This meant their method needed prior information about the category of object with which the robot interacted. Also, in their method, they make a single prediction before interaction and then rely on controller robustness to account for mistaken predictions.

A closed-loop learning estimation method was proposed by \citet{Schiavi2022}. This method estimates articulation affordance from vision at multiple time steps during the interaction. A sampling-based controller solves for the optimal opening trajectory. When opening, the object becomes stagnant due to torque limits, the robot releases the object and moves to a configuration to view the full object again, then makes a new vision-based estimate of the articulation. The requirement to let go of the object to re-view it, slows the opening down, and assumes the object's door will not snap back shut or fall open when released. These approaches have relied heavily on robust and compliant controllers to account for any errors in articulation estimation. In contrast, our work updates the estimation of the articulation model seamlessly during interaction, enabling the use of much simpler methods for motion generation and control.

\section{Screw Theory Background}
\label{sec:background}
Screw theory is the geometric interpretation of twists that can be used to represent any rigid body motion (Chasles theorem)~\cite{Murray1994}. Screw motions are parameterized by the twist $\xi = ( \linvel, \rotvel ), \text{where} ~ \linvel, \rotvel \in \mathbb{R}^3$. The variable $\linvel$ represents the linear motion and $\rotvel$ the rotation.
We can convert this to a tangent space to $\SEthree$ using $\hat{\xi}$ as
\begin{equation}
	\label{eq:xi_hat}
	\hat{\xi} = \begin{bmatrix}
	\hat{\rotvel} & \linvel\\
	0 & 0
\end{bmatrix} \in \mathfrak{se}(3),
\end{equation}
where the hat operator $\hat{(\cdot)}$ is defined as:
\begin{equation}
	\label{eq:hat}
	\hat{\boldsymbol{\omega}} = \begin{bmatrix}
	0 & -\omega_z & \omega_y\\
	\omega_z & 0 & -\omega_x\\
	-\omega_y & \omega_x & 0
\end{bmatrix}.
\end{equation}
In screw theory, $\xi$ is a parametrization of motion direction and $\theta$ is a signed scalar representing the motion amount. In the pure rotation case, $\theta$ has the units of radians, and in the pure translation case, meters. The tangent space \eqref{eq:xi_hat} can be converted to the homogeneous transformation $\mathbf{T}(\hat{\xi}, \theta) \in \SEthree$ using the exponential map $\exp: \mathfrak{se}(3) \to \SEthree$ from \citet{Murray1994} where
\begin{align}
    \exp{(\hat{\xi}\theta)} &=\begin{bmatrix}
\exp{(\hat{\rotvel}\theta)} & (\Identity - \exp{(\hat{\rotvel}\theta)})(\rotvel \times \linvel) + \rotvel\rotvel^T\linvel\theta \\
0 & 1
\end{bmatrix},
\end{align}
and $\exp{(\hat{\rotvel}\theta)}$ is solved by using the Rodriguez Formula:
\begin{align}
        \exp{(\hat{\rotvel}\theta)} &= \Identity + \hat{\rotvel}\theta + \frac{(\hat{\rotvel}\theta)^2}{2!} + \frac{(\hat{\rotvel}\theta)^3}{3!} + ...\\
        &= \Identity + \hat{\rotvel}\sin{\theta} + \rotvel^2(1-\cos{\theta}).
\end{align}

If we define a fixed world frame $\World$, then the frame attached to the moving part of an articulated object (e.g., the cabinet door) is given the frame $\mathtt{A}$ and the homogeneous transform between them is given as $\mathbf{T}_\mathtt{WA} \in \SEthree$.
The other non-moving part of the object (e.g., the cabinet base) is given the frame $\mathtt{B}$ and its pose in $\World$ is defined as $\mathbf{T}_\mathtt{WB} \in \SEthree$. Since $\exp{(\hat{\rotvel}\theta)} \in \SEthree$ defines the homogeneous transform from the object base $\mathtt{B}$ and articulated part $\mathtt{A}$, we can express the screw transform as:
\begin{align}
    \label{eq:expmap}
    \mathbf{T}_{\mathtt{BA}}(\hat{\xi}, \theta) = \exp{(\xihat\theta)},
\end{align}
where $\theta \in \mathbb{R}$ is the articulation configuration. The two object parts are then connected by
\begin{align}
\label{eq:transform}
    \mathbf{T}_\mathtt{WA} = \mathbf{T}_\mathtt{WB}\partBA(\hat{\xi}, \theta).
\end{align}

\section{Preliminaries}
The goal of this work is to estimate online the Maximum-A-Posteriori (MAP) state of a single joint from visual and proprioceptive sensing. We define the state $\State(t)$ at time $t$ as
\begin{equation}
\State(t) \coloneqq
\left[\xi,\theta(t) \right] \in \Real^{7},
\end{equation}
where $\xi$ are the screw parameters which we assume to be constant for all time and $\theta(t)$ is the configuration of articulation at time $t$. We assume the object is composed of only two parts connected by a single joint. This encompasses the vast majority of articulated objects and therefore is a reasonable simplification. We define the pose of each part in the world frame $\World$ as $\mathbf{T_{\mathtt{WA}}}, \mathbf{T_{\mathtt{WB}}} \in \SEthree$ where $\mathbf{T_{\mathtt{WB}}}$ is the \textit{base} part that is static and $\mathbf{T_{\mathtt{WA}}}$ is the \textit{articulated} part which the robot grasps, e.g. the door. 

We jointly estimate $\mathsf{K}$ states and part poses with time indices $k$; so that the set of all estimated states and articulated part poses can be written as $\calX = \{\State_k, \partA_{k}, \partB_{k}\}_{k \in \mathsf{K}}$, dropping world reference frames for brevity.
Our method fuses measurements from three sources: point cloud measurements from an initial visual inspection, force measurements from a wrist-mounted 6-axis force/torque sensor, and kinematic measurements from joint encoders in the robot's arm. We use $\mathsf{P}$ point clouds, which are each associated with a prediction on $\xi$. Without loss of generality, we set $\mathsf{P} = 1$ with one visual measurement at the beginning. In future work,
we seek to add multiple vision-based predictions that can be added to the factor graph asynchronously during manipulation. Force measurements are added at time indices $f$ up to a maximum of $\mathsf{F}$ measurements. We use $\mathsf{K}$ kinematic measurements at times $k$, which are each associated with a state estimate. The times $k$ are only selected while the robot is in contact with the object and after the articulated part has been moved a certain distance $d$ to avoid taking too many measurements. 

Finally, the set of all measurements are then grouped as $\calZ = \{\calP, \calF_{f}, \calK_{k}\}_{f \in \mathsf{F} k \in \mathsf{K}}$ where $\calP$ is the point cloud measurement, $\calF$ are the force measurements and $\calK$ the pose measurements from the robot's forward kinematics. 

\section{Factor Graph Formulation}
We maximize the likelihood  of the measurements $\calZ$, given the history of states $\calX$:
\begin{equation}
	\calX^* = \argmax_{\calX} p(\calX|\calZ) \propto p(\State_0)p(\calZ|\calX),
	\label{eq:posterior}
\end{equation}
where $\calX^*$ is our MAP estimate of the articulation.

\begin{figure}[t]
	\centering
	\includegraphics[width=\columnwidth]{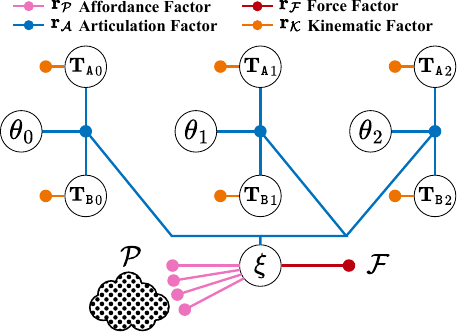}
	\caption{The factor graph shows the variables we are estimating: $\partA(t), \partB(t), \theta(t)$ and $\xi$, which exists at only one time step in the factor graph. We show three time steps, including the initial visual affordance factor, which provides a prior estimate on $\xi$ as a unary factor.}
	\label{fig:fg}
\end{figure}

We assume the measurements are conditionally independent and corrupted by zero-mean Gaussian noise. Therefore, \eqref{eq:posterior} can be expressed as the following least squares minimization:
\begin{align}
\begin{split}
	\calX^{*} = \argmin_{\calX} \|\mathbf{r}_\textup{0}\|^2_{\Sigma_\textup{0}}
	               &+  \|\mathbf{r}_{\calP}\|^2_{\Sigma_{\calP}}
                   + \sum_{f \in \mathsf{F}} \|\mathbf{r}_{\calF_{f}}\|^2_{\Sigma_{\calF}}\\
	               &+ \sum_{k \in \mathsf{K}} \Big( \|\mathbf{r}_{\calA_{k}} \|^2_{\Sigma_{\calA}} + \|\mathbf{r}_{\calK_{k}} \|^2_{\Sigma_{\calK}} \Big),
	\label{eq:cost-function}
\end{split}
\end{align}
where each term is a residual $\mathbf{r}$ associated with a measurement type and assumed to be corrupted by zero-mean Gaussian noise with covariance according to the measurement. A factor graph can be used to graphically represent \eqref{eq:cost-function} as shown in Fig~\ref{fig:fg}, where large white circles represent the variables we would like to estimate and the smaller colored circles represent the residuals as factors. The implementation of the factors is detailed in the following section.

\section{Method}
\label{sec:method}
In this section, we describe how the three type of measurement (point cloud $\calP$, force $\calF$ and kinematics $\calK$) are fused together in our factor graph, using four factors (Affordance $\mathbf{r}_{\calP}$, Articulation $\mathbf{r}_{\calA}$, Force $\mathbf{r}_{\calF}$ and Kinematics $\mathbf{r}_{\calK}$) to estimate articulation online.

\subsection{Uncertainty-aware Articulation Prediction from Vision}
\label{sec:uncerteinty-prediction}

\begin{figure}[t]
	\centering
	\includegraphics[width=\columnwidth]{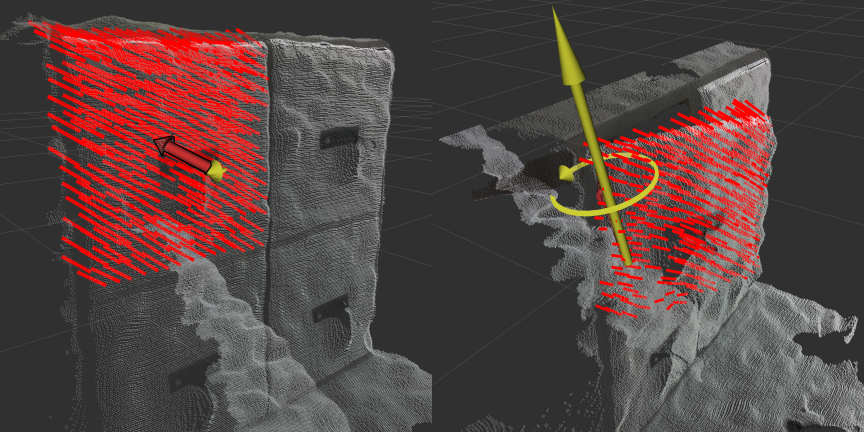}
	\caption{Example affordance predictions from the neural network from \citet{Buchanan2024}: prismatic left and revolute right. The small red lines are the output of the network, predicting articulation flow on the segmented points. The large red and yellow arrows indicated the resulting joint prediction from plane fitting as was done in \citet{Buchanan2024}.}
	\label{fig:network-output}
\end{figure}

We are interested in using a deep neural network to predict articulation affordance from visual measurements.
This affordance can be parameterized as a point cloud whereby each 3-dimensional point encodes a normalized, instantaneous velocity of that point given a small amount of articulation.
As shown in Fig.~\ref{fig:network-output}, for a prismatic joint, all vectors will point along the axis of motion equally.
For a revolute joint, all vectors will point tangent to the circular trajectory, with the vectors further from the axis of rotation longer. 
\citet{Zeng2021} first introduced this representation of articulation affordance, describing it as a motion residual \emph{flow}.
Later, \citet{Eisner2022,Eisner2023} improved the implementation with Flowbot3D and \citet{Buchanan2024} used the network from Flowbot3D in their framework. 

We introduce a new neural network which, like Flowbot3d, uses PointNet++~\cite{Charles2017} as the underlying architecture. Our neural network takes in a point cloud $\calP$ consisting of $\mathsf{N}$ points where $n \in \mathsf{N}$ and $\textbf{p}_n \in \Real^4$ which encode 3-dimensional position and a mask indicating whether the point belongs to the articulated part or the fixed-base part. The network then predicts flow $\hat{\flow}_n \in \Real^3$ for each point on the articulated part. 

In our previous work~\cite{Buchanan2024} we used the Mean Squared Error (MSE) loss function to train the network to only predict flow:
\begin{align}
    \calL_{\text{MSE}}(\flow, \hat{\flow}) = \frac{1}{p}\sum_{i=1}^{p}\norm{\flow_i - \hat{\flow}_i}^2,
\end{align}
In this work, we seek to train the neural network to also predict its \emph{aleatoric} uncertainty for each point-wise prediction of flow. To achieve this, we use change the loss function to  the method shown in \citet{Russell2021} to change the loss function to the following Gaussian Maximum Likelihood (ML) loss:
\begin{align}
\begin{split}
    &\calL_{\text{ML}}(\flow, \hat{\Cov}, \hat{\flow}) \\
    &= \frac{1}{N}\sum_{n=1}^{N}-\text{log} \left( \frac{1}{\sqrt{8\pi^3\text{det}(\hat{\Cov}_n)}} e^{-\frac{1}{2}\norm{\flow_n - \hat{\flow}_n}_{\hat{\Cov}_n}^{2}} \right).
\end{split}
\end{align}
This enables the network to learn to predict articulation flow in a supervised manner from the labels $\flow$, and learn the uncertainty $\hat{\Cov}$ in an unsupervised manner. The network function $f$ with trained weights $\Theta$ can then be represented as:
\begin{align}
    f_\Theta  : \left(\calP \right) \mapsto (\hat{\flow}_i,
    \hat{\mathbf{u}}_i),
\end{align}
where $\hat{\mathbf{u}}_i \in \Real^3$, can be formulated into a covariance matrix using 
\begin{align}
    \hat{\Cov}_i(\hat{\mathbf{u}}_i) = \text{diag}(e^{2\hat{u}_x}, e^{2\hat{u}_y}, e^{2\hat{u}_z}).
    \label{eq:pred-cov}
\end{align}

\begin{figure*}[t]
	\centering
	\includegraphics[width=17.5cm]{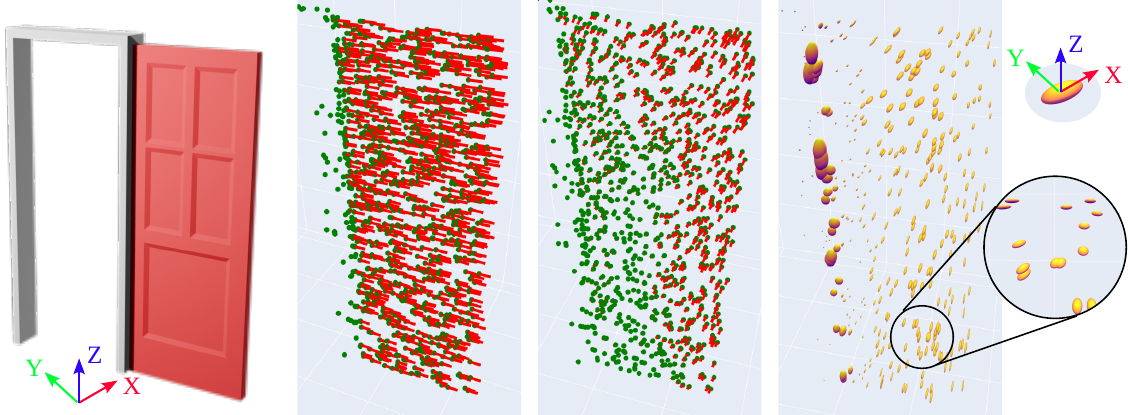}
	\caption{Example output of articulation flow prediction with covariances. \textbf{Left}: rendering of simulated sliding door, note the axis. \textbf{Center left}: green point cloud measurement of the door with the ground truth articulation flow shown as red lines. \textbf{Center right}: predicted articulation flow shown as red lines. The neural network has mistaken the door for revolute with a joint on the right side of the door frame. \textbf{Right}: the covariance for each articulation flow vector. There is the highest covariance in the x direction, showing a high degree of uncertainty with this articulation. The next highest uncertainty is in the y direction, followed by z.}
	\label{fig:covariance-example}
\end{figure*}

An example output of this new articulation prediction is shown in Fig.~\ref{fig:covariance-example}. In this simulated sliding door example, the door is almost fully open, which makes the articulation visually ambiguous. The prediction of the network shows a belief that the door is revolute about the door frame. However, inspection of the covariance shows there is the largest uncertainty in the $x$ direction, followed by $y$, with the lowest uncertainty in the $z$ direction. This will be useful later when we introduce the force factor, which will "correct" motion in the $x$ direction to be zero with very low uncertainty, allowing the articulation estimation to collapse to the $y$ direction.

\subsection{New Affordance Factor}
\label{sec:new-affordance-factor}
In previous work on articulation estimation from deep learning affordance predictions, a hand-crafted approach was used to incorporate flow predictions into the factor graph~\cite{Buchanan2024}. This involved fitting two planes to the initial point cloud and to the point cloud representing a small articulation.
The intersection of these planes represented a measurement on a revolute joint.
If the intersection was very far away, then the direction of flow was used as a measurement on a prismatic joint.
These articulation predictions $\hat{\xi}$ were used directly on the articulation estimate as unary factors with the following residual:
\begin{align}
    \mathbf{r}_{\calP} = \xi - \hat{\xi}.
\end{align}
This required a hand-tuned uncertainty ($\sigma_{\calP} = 1e^{-3}$) which did not capture the true uncertainty of the neural network. 

In our work, we instead introduce a new affordance factor which directly integrates the  predicted uncertainty $\hat{\Cov}$. First, we change the single articulation factor from \citet{Buchanan2024} to instead be a sum of per-point factors for each of the $\mathsf{N}$ points in the point cloud $\calP$:
\begin{align}
\label{eq:sum-art}
\begin{split}
    \mathbf{r}_{\calP} &= \sum_{i \in N} \mathbf{r}_{\calP_{i}},
    \end{split}
\end{align}
As discussed in Sec.~\ref{sec:uncerteinty-prediction}, each flow vector represents a position change of a point lying on the moving part of the object as a result of a small change in articulation angle: $\theta = 0.05$ (this $\theta$ increment is also used for generating training data in simulation). The new point position after the articulation can be written as: 
\begin{align}
\label{eq:flow-move}
    \hat{\mathbf{p}}^{+}_{i} = \hat{\flow}_i + \mathbf{p}_i.
\end{align}
Equivalently, using the equation for articulation homogeneous transform\footnote{In this case, $\mathbf{p}_i$ and $\mathbf{p}_i^{+}$ are represented in homogeneous coordinates. We convert $\mathbf{p}_i$ back to Cartesian coordinates later.}, we can write:
\begin{align}
\label{eq:flow-transform}
    \mathbf{p}^{+}_{i} = \partBA(\hat{\xi}, \theta)\mathbf{p}_i.
\end{align}
If we set $\theta$ to be very small ($\approx0.05$), then we can expect $\hat{\mathbf{p}}^{+}_{i}$ from \eqref{eq:flow-move} to be equivalent to $\mathbf{p}^{+}_{i}$ from \eqref{eq:flow-transform}, and therefore we can use the following residual on a per-point basis:
\begin{align}
\begin{split}
    \mathbf{r}_{\calP_{i}} &=\mathbf{p}^{+}_{i} - \hat{\mathbf{p}}^{+}_{i}\\
        &= \partBA(\hat{\xi}, \theta)\mathbf{p}_i - \hat{\flow}_i - \mathbf{p}_i,
        \label{eq:art-residual}
    \end{split}
\end{align}
which is conditioned on the predicted covariance $\hat{\Cov}_i$ from the neural network in \eqref{eq:pred-cov}. Therefore, the point cloud articulation residual $\mathbf{r}_{\calP}$ in \eqref{eq:cost-function} is replaced with a sum of per-point residuals (\eqref{eq:art-residual} and \eqref{eq:sum-art}) and is visually represented in Fig.~\ref{fig:fg}.

\subsection{Force Factor}
\label{sec:force-factor}
Another limitation of previous works in this area was the requirement for the initial opening direction to be pulling away from the object. This meant revolute doors or prismatic drawers could be estimated, but not prismatic sliding doors such as the bottom right door of the cabinet shown in Fig.~\ref{fig:motivation-cabinet}.
This was because the neural network would make a prediction of a drawer-like prismatic joint, and the robot would pull backwards on the door.
However, because the door would not move, there was no opportunity to collect kinematic measurements and update the articulation estimate.

As a solution in this work, we propose using force measurements from a wrist-mounted force sensor to infer articulation.
We use force measurements at the beginning of the interaction, after grasping the articulated part, but before any motion.
If the reaction force measurement reaches a given threshold when attempting to open the door, then the factor graph incorporates this force~$\hat{\mathbf{F}}$ as an additional factor when solving for the new estimate of the articulation.
Although force and torque can be used to guide a robot controller to minimize torque during opening, it is less straightforward to use these measurements to infer the articulation parameters.
This is because a reaction force/torque measurement only informs that a particular direction is not a valid motion, rather than informing which alternative motion would be correct.

\begin{figure}[t]
	\centering
	\includegraphics[width=0.7\columnwidth]{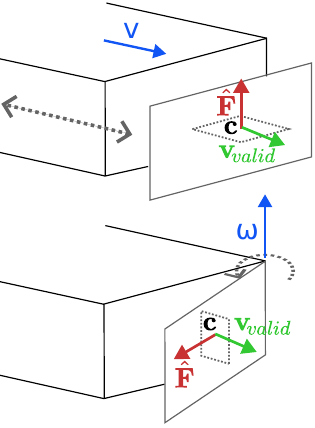}
	\caption{Example of relationship between applied direction of motion (grey), measured reaction force, and valid direction of motion. \textbf{Top}: a downward force is applied to a prismatic joint, which results in the upward reaction force $\hat{\mathbf{F}}$. This is orthonormal to a plane (gray dotted plane) on which we know the valid motion $\linvel_{valid}$ must lie. \textbf{Bottom}: a force is applied towards the hinge of a revolute joint, resulting in a reaction force perpendicular to the direction of motion and orthonormal to a plane on which $\linvel_{valid}$ lies.}
	\label{fig:force-example}
\end{figure}

However, we can still use this information about invalid motion to rule out possible articulations. If a robot attempts to open an articulated object, and there is no motion, then the direction of force must be orthogonal to the valid direction of motion. This can be viewed as a simplified version of the approach used in \citet{MartinMartin2022} in which we do not require a particle filter.

The direction of force can be expressed as:
\begin{align}
    \label{eq:force}
    \linvel_{valid} \cdot \hat{\mathbf{F}}  = 0,
\end{align}
where $\hat{\mathbf{F}} \in \Real^3$ is the measured reaction force measurement and $\linvel_{valid} = \linvel + \rotvel \times \vec{c}$ defines the true, valid instantaneous direction of motion for a point $\vec{c}$ on the articulated part. This relationship is demonstrated for both prismatic and revolute joints in Fig.~\ref{fig:force-example}.

Intuitively, this means if a robot attempts to pull backward on a sliding door, it can be inferred that the door only opens in a direction that spans the vertical plane (i.e., left/right or up/down). For this to hold, we make a few assumptions:
\begin{itemize}
    \item If the articulation is a revolute joint, the grasp point is not on the hinge. We can make this assumption because a human operator is providing the grasp point to the robot.
    \item While applying the force, the articulated object does not move. 
\end{itemize}

If these two assumptions hold, we can use \eqref{eq:force} to correct the estimated direction of motion $\linvel_{est}$ so that it lies on a plane with normal $\hat{\mathbf{F}}$. To perform the rotation, we first find a vector orthogonal to both $\linvel_{est}$ and $\hat{\mathbf{F}}$:
\begin{equation}
    \vec{t} = \linvel_{est} \times \hat{\mathbf{F}}.
\end{equation}
If $\vec{t}$ is a zero vector (i.e., $\linvel_{est}$ is parallel or anti-parallel to $\hat{\mathbf{F}}$), we select an alternative perpendicular vector by taking the cross product of $\hat{\mathbf{F}}$ with a standard basis vector while ensuring:
\begin{equation}
    \hat{\vec{t}} = \frac{\vec{t}}{\|\vec{t}\|}.
\end{equation}
This guarantees $\hat{\vec{t}}$ lies in the plane and is orthogonal to both $\linvel_{est}$ and $\hat{\mathbf{F}}$.
The vector is then rotated by $90^\circ$ using the cross product:
\begin{equation}
    \vec{v}_{\text{rot}} = \hat{t} \times \linvel_{est}
\end{equation}
Therefore, we can then define a residual as:
\begin{align}
    \label{eq:force-residual}
    \mathbf{r}_{\calF_{f}} &= \linvel_{est} - \vec{v}_{\text{rot}}
\end{align}

This will push the estimate of $\linvel_{est} = \linvel + \rotvel \times \vec{c}$ onto the plane where a possible direction of motion exists. When used with the affordance factors as described in Sec.~\ref{sec:new-affordance-factor}, this will push the optimized result towards the next most likely articulation with a motion that lies on the plane.
For the force factor, we hand-tune the uncertainty to be very low ($\Sigma_{\calF}=1e^{-6}$) so that a single factor can correct for the affordance factors.

\subsection{Kinematic Factor}
\label{sec:kinematic-factor}
We optimize for both the articulation state $\State$ and the part poses $\partA$ and $\partB$. At time $k$, the forward kinematics of the robot are used to compute the end-effector pose, which is assumed to have a rigid grasp of the articulated part of the object. This provides measurements on $\partA$ during interaction. Additionally, we assume $\partB$ does not move and therefore, we reuse the initial grasp pose at every time $k$. To account for a small amount of slippage, we associate an uncertainty with these measurements, and the value is manually tuned $\sigma_\calK =1e^{-3}$.  The residual $\mathbf{r}_{\calK_{k}}$ is the default $\SEthree$ unary factor in GTSAM~\cite{gtsam}.

\subsection{Articulation Factor}
\label{sec:art-factor}
The fourth and final factor we use is equivalent to the articulation factor as used in previous work~\cite{Buchanan2024}. This factor connects the variables $\State$, $\partA$ and $\partB$ in the factor graph using the articulation screw model explained in Sec.~\ref{sec:background}. We compare the estimated part poses to the expected articulation model in (Sec.~\ref{eq:transform}). As in \citet{Buchanan2024}, putting these together gives us the articulation residual as:
\begin{align}
    \mathbf{r}_{\calA_{k}} = \partBA(\xihat, \theta_k) \boxminus {\partB^{-1}_k} \partA_k,
\end{align}
where $\boxminus$ is a pose differencing over the manifold using the logarithm map:
\begin{align}
    \partA \boxminus \partB = \text{Log}({\partB_k}^{-1} \partA_k) \in \mathfrak{so}(3).
\end{align}

\section{Implementation}
\label{sec:implementation}
\begin{figure*}[t!]
	\centering
	\includegraphics[width=0.9\linewidth]{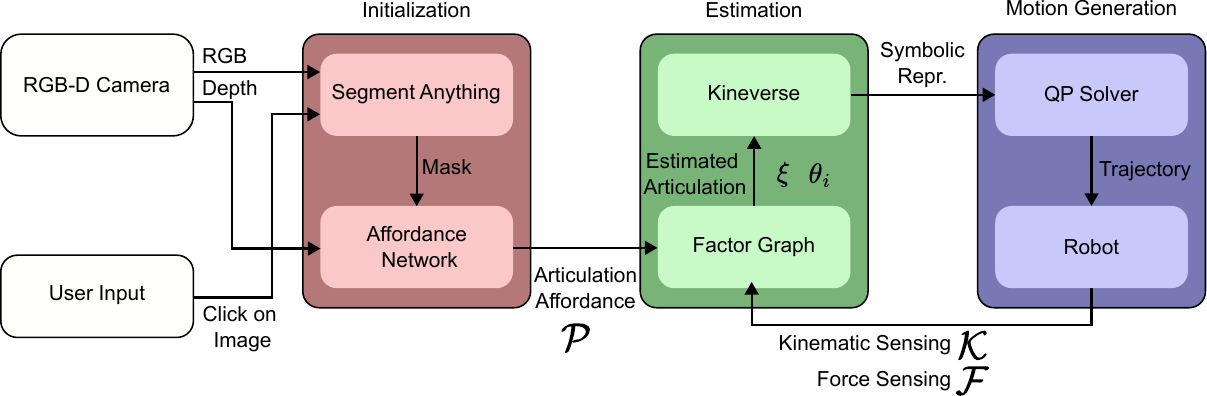}
	\caption{Full system with information flow. An RGB-D camera provides RGB images, which are segmented with the click prompt from a human user. This generates a mask on the articulated part, which with depth information from the camera, predicts initial articulation parameters. This is provided to the factor graph, which also uses kinematic measurements of the end effector to estimate the object articulation. The estimated articulation updates the symbolic math representation of both robot and object, which is then formulated as a quadratic programming (QP) problem to solve for the robot trajectory.}
	\label{fig:pipeline}
\end{figure*}

This section describes the full system implementation for shared autonomy as shown in Fig.~\ref{fig:pipeline}. The system consists of three modules: Initialization, which occurs once at the beginning; Estimation, which runs online to estimate the articulation; and Motion Generation, which computes the robot's trajectory to open the object. Estimation and Motion Generation both run online, generating a new trajectory for each new articulation estimate.

\subsection{Initialization}
For the initialization module, we use the latest advances in deep learning for articulated objects and introduce a system of shared autonomy. First, a user is presented with a video feed of the object and clicks on the desired grasp point. With this query point, we use the publically available segmentation tool Segment Anything (SAM)~\cite{Kirillov2023} to segment a mask of the non-static part. We make use of the open-source ROS wrapper for SAM first presented in \citet{Buchanan2024}.
The image mask and associated point cloud are then passed to the network, which predicts the articulation affordance for each masked point as described in Sec.~\ref{sec:uncerteinty-prediction}. The neural network is trained from examples of articulated objects in PyBullet simulation using the PartNet-Mobility dataset~\cite{Mo2018}. Therefore, the output of the initialization step is the point cloud affordance $\calP$, and the 3D point associated with the user's click, which will be used as the first planning goal for the robot.

\subsection{Estimation}

Once the system has been initialized, the factor graph is first optimized using the affordance prediction $\calP$. We use GTSAM~\cite{gtsam} for the implementation of the factor graph. The first optimization results in the first prediction of articulation, which is immediately sent to the Motion Generation module. If the door that the robot is interacting with begins to move, kinematic measurements are then added to the factor graph for every distance $d$ the end-effector moves.
We use the Kineverse articulation model framework~\cite{Rofer2022} for representing both the robot and the articulated object forward kinematics and constraints. Kineverse uses the CasADi symbolic math back-end~\cite{Andersson2019}, enabling effortless computation of gradients for arbitrary expressions, such as articulations.

A major limitation with previous work was that if the initial prediction of articulation was orthogonal to the actual articulation, then the end-effector would not move during interaction, and therefore, the estimation could not be updated from kinematic measurements. For example, in the bottom right drawer in Fig.~\ref{fig:motivation-cabinet}, the door slides open to the right. However, the network usually predicted a prismatic joint pulling backwards. The robot arm would pull on the door handle and not move, thereby learning nothing new about the articulation.
In this work, we are able to overcome this challenge using the new articulation factor as described in Sec.~\ref{sec:uncerteinty-prediction} in conjunction with force factor as described in Sec.~\ref{sec:force-factor}. The robot has a force/torque sensor in its wrist. As it attempts to open the door, if the reaction force is larger than a set threshold and the door has not moved, this triggers the addition of a force factor and an additional optimization. This results in a new articulation estimation, which is passed to the Motion Generation module. In this way, the robot will continuously attempt to open the door and use the reaction force to guide the next estimate.

\subsection{Motion Generation}
\label{sec:motion-generation}
This module computes the desired robot configurations~$\vec{q}\in\Real^7$, to open the object, given the latest estimate of the articulation~$\xi$. We define the robot end-effector frame as $\EndEff$ and model the forward kinematics of the robot end-effector as~$\mat{T}_{\World\EndEff}(\vec{q})$. 

To compute the goal forward kinematics for opening the articulated object, we slightly rewrite~\eqref{eq:expmap} as
\begin{align}
\partWA(\xihat, \theta_t) &= \partWB \cdot \partBA(\xihat, \theta_t),
\end{align}
with~$\partBA(\xihat, \theta_t)$ provided from the latest estimate of $\xi$ and using a goal $\theta_t$. The pose $\partWB$ is a static transformation composed of the user defined grasp position~$\vec{p}_{\World\mathtt{B}}$ and a predetermined grasp orientation~$\vec{R}_{\World\mathtt{B}}$.

Once the robot grasps the object handle, we set $\theta_0 = 0$, which leads to $\partBA(\xihat, 0) = \Identity_{4\times4}$.
We then progressively increment the desired articulation configuration $\theta_{t + 1} = \theta_t + gv \Delta t$, with $gv$ being a constant speed for opening/closing the articulation, up to the articulation limit after which we invert the sign of~$gv$.
For each $\theta_t$, and given an estimate of $\xi$, we solve the inverse kinematics (IK) problem, subject to the condition~$\mat{T}_{\World\EndEff}(\vec{q}_{t+1}) = \partWA(\xihat, \theta_t)$.
More specifically, we define the IK problem as a non-linear optimization problem where we encode the following task space constraints
\begin{align}
\begin{split}
    \norm{\mat{p}_{\World\EndEff}(\vec{q}_{t+1}) - \mat{p}_{\World\Articulation}(\xihat, \theta_t)}_F^2 &= 0\\
    \norm{\mat{R}_{\World\EndEff}(\vec{q}_{t+1}) - \mat{R}_{\World\Articulation}(\xihat, \theta_t)}_F^2 &= 0
\end{split}
\label{eq:pose_constraint}
\end{align}
where~$\norm{\cdot}_F$ denotes a Frobenius norm.

We exploit the differentiability of the constraints in~\eqref{eq:pose_constraint} w.r.t.~to \vec{q}, to linearize the problem, and solve it sequentially until constraint satisfaction as a quadratic program (QP):
\begin{equation}
    \begin{aligned}
        \argmin_{\mathbf{x}}\,\frac{1}{2}\mathbf{x}^T\mathbf{C}\mathbf{x} \quad
        \textrm{s.t.} \quad & \mathbf{lb}   \leq \mathbf{x}  \leq \mathbf{ub} \\
                        & \mathbf{lb}_A \leq \mathbf{Ax} \leq \mathbf{ub}_A,
    \end{aligned}
    \label{eq:qp}
\end{equation}
where $\vec{x} = \langle \vec{\fd{q}}, \vec{s}\rangle$ is a vector of joint velocities and slack variables $\vec{s}$, and $\mat{A}$ is the Jacobian of the task constraints and the associated slack variables.
\eqref{eq:qp} also encodes bounds on robot joint positions and velocities. We use our Kineverse~\cite{Rofer2022} symbolic representation for computing the Jacobians, as well as encoding and solving the problem in~\eqref{eq:qp}.

Finally, we command the resulting joint positions $\vec{q}_{t+1}$ to the robot in compliant mode.
Therefore, if the articulation estimation~$\xi$ is inaccurate, the robot can comply with the physical articulation, leading to an end-effector pose that is different from~$\mat{T}_{\World\EndEff}(\vec{q}_{t+1})$. The actual end-effector pose~$\partWA(\xihat, \theta_{t+1})$ is added to the graph as a measurement on $\partWA$.

\section{Experiments}
\label{sec:experiments}
In the following section, we describe the experiments we conducted to evaluate our method, and we report the results. Discussion of the results follows in Sec.~\ref{sec:discussion}. 

\begin{figure*}[t]
	\centering
	\includegraphics[width=\linewidth]{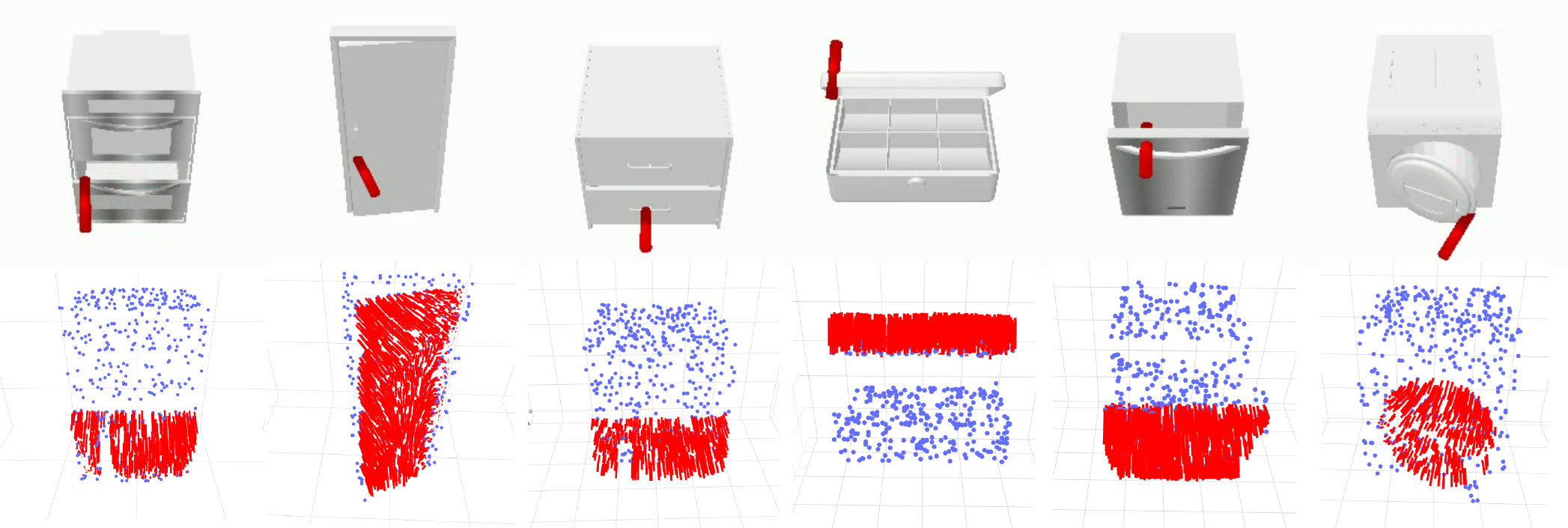}
	\caption{Example output of simulation experiments. \textbf{Top Row:} Objects from the PartNet Mobility dataset are rendered in PyBullet. Red lines indicate the pulling direction resulting from the articulation estimate. \textbf{Bottom Row:} The input point cloud is shown in blue, and the output articulation flow predicted from our neural network is overlaid as red lines.}
	\label{fig:simulation-example}
\end{figure*}

\subsection{Simulation Experiments}
\label{sec:simulation}
In these initial experiments, we compare our uncertainty-aware articulation prediction method to the affordance prediction method of Flowbot3D~\cite{Eisner2022}. We simulated point cloud data in PyBullet for several unseen articulated objects from the PartNet-Mobility dataset, and then each neural network made predictions on the articulation of the object. For each method, we selected the grasp point in the same way as the authors of Flowbot3D, which is to select the point with the largest magnitude of predicted flow. The articulation is then predicted using each method, and an applied force is simulated on the object, at the grasp point, in the direction of articulation opening. Each object starts closed, and we consider an opening to be successful if the object has been opened 90\% of its limit, which was the same criteria as \citet{Eisner2022}.

We chose to simulate an applied force rather than use a floating end-effector because we found that the end-effector often was able to open objects using unrealistic methods, such as passing through the object and then opening it from the inside. Additionally, due to poorly modeled contact physics, the end-effector would sometimes experience unrealistically large forces, causing the grasp to slip.

We compared the case where a single articulation prediction is made at the beginning (Single), and where continuous point cloud measurements are simulated and articulation is continuously predicted during the interaction (Multi). In the Multi experiments, each new articulation prediction updates the pulling direction (but not the grasp point). This way, if the first prediction was sufficient to open the object a small amount, but not fully, additional predictions can take advantage of the slight opening to make better predictions.

We also compare our method with different numbers of articulation factors. We subsample the point cloud to 200, 500, and 1000 points. Each point results in an additional factor in the factor graph, which increases the time for optimization.

\subsubsection{Results}
The results are summarized in Tab.~\ref{tab:results}, and some example experiments are shown in Fig.~\ref{fig:simulation-example}. It is clear that Multi inference is significantly superior to Single as it can update the prediction during interaction. However, for several reasons we believe that Multi inference is not realistic for deployment on robot hardware. Firstly, if the camera is installed on the robot end-effector, it would not be possible to observe the articulated object during interaction and would instead require the robot to let go of the object and re-observe the scene as in \citet{Schiavi2022}. If, on the other hand, the camera is installed externally or on another part of the robot, such as a humanoid robot's head, there would still be the issue of occlusion due to the interacting robot arm, and the segmentation mask would need continuous updating while the articulated part is moved.

Instead, we believe the Single inference approach is more realistic when combined with the proposed kinematic and force-based sensing. Additionally, when comparing inference times in Tab.~\ref{tab:results}, we see a significant increase in latency as more factors are added to the factor graph. Because we intend to only make a single inference at the beginning of interaction, this increase is acceptable.
Finally, we note that our method with 500 and 1000 factors outperforms Flowbot3D by 8.7\% and 11.2\% respectively. This is because our approach integrates a large number of articulation points to optimize an overall solution for articulation. In contrast, Flowbot3D selects the single largest point. While this allows their method to have very little latency, it increases variability in predictions, which leads to an overall lower success rate. In our method, increasing the number of articulation factors improves performance; however, we noted no further improvement beyond 1000 points, and we used this value for later real robot experiments.

\begin{table}[t!]
\centering
\caption{Simulation results presented as percent success and time for each inference in seconds.}
\label{tab:results}
\resizebox{\linewidth}{!}{
\begin{tabular}{ccccc}
\toprule
\textbf{Method} & \textbf{Single} & \textbf{Multi} & \textbf{Average} & \textbf{Worst}\\
\midrule
Flowbot 3D~\cite{Eisner2022} & 53.67\% & 61.47\% & 0.01\,s& 0.02\,s\\
Art. Factor 200 & 51.74\% & 61.49\% & 0.03\,s& 0.35\,s\\
Art. Factor 500 & 56.60\% & 66.81\% & 0.11\,s & 0.82\,s\\
Art. Factor 1000 & 57.71\% & 68.37\% & 0.22\,s & 1.55\,s\\
\bottomrule
\end{tabular}
}\\
\end{table}

\subsection{Hand Guiding Experiments}
\label{sec:hand-guiding}
In these experiments, we investigated the accuracy of our kinematics-based articulation estimation. We compared our method against \citet{Heppert2022} which also uses factor graphs to estimate a screw parameterization. The authors kindly granted us access to their code for direct comparison.

\begin{figure}[t]
	\centering
	\includegraphics[width=\columnwidth]{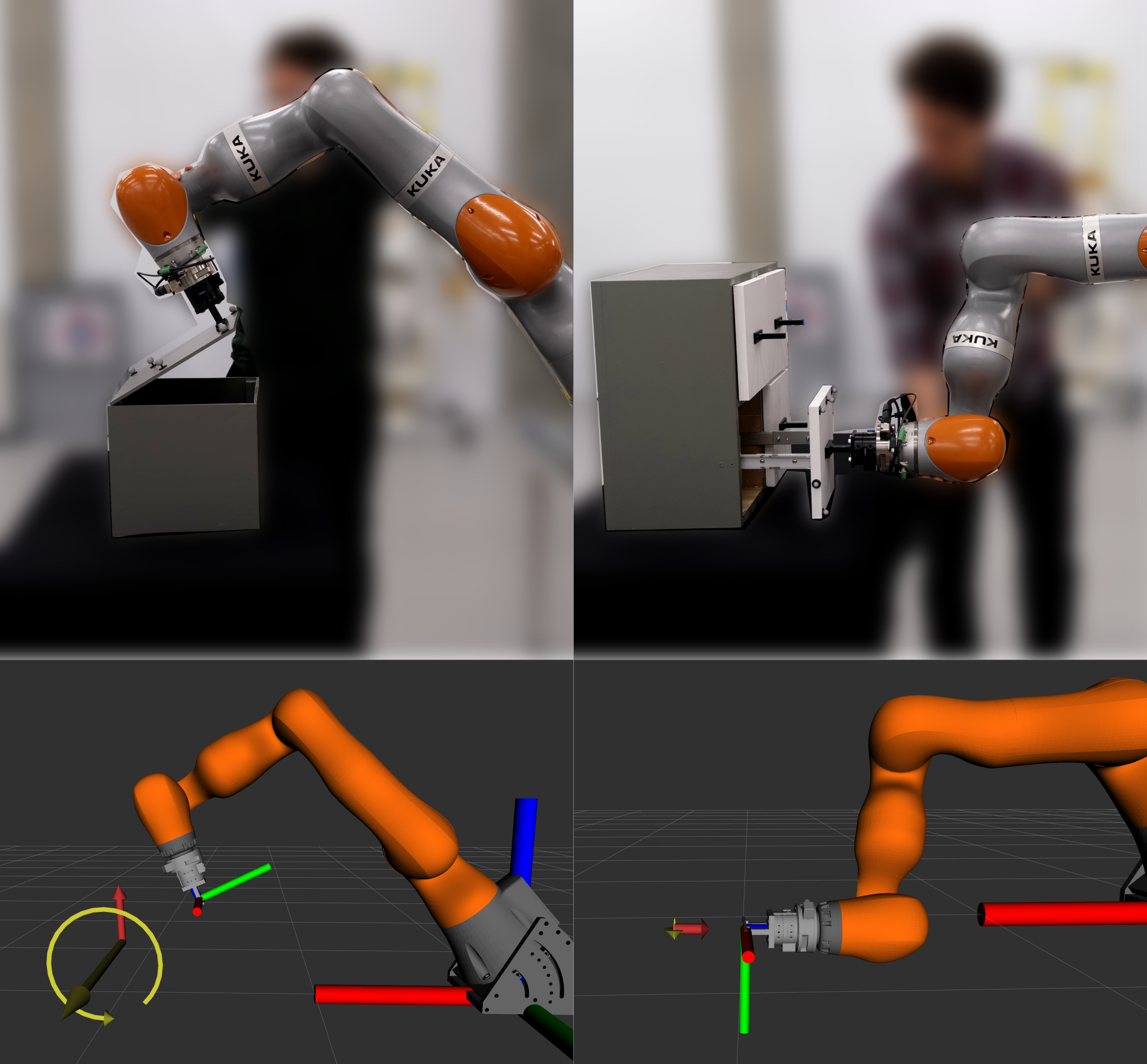}
	\caption{\textbf{Top}: hand guiding experiments for revolute (left) and prismatic (right) joints. Motion capture markers are only for ground truth reference. \textbf{Bottom}: the resulting estimated articulation from joint encoder sensors. Yellow arrows show $\rotvel$ while red arrows show $\linvel$. The large axis is the base frame of the robot which is used for $\World$ while the small axis is the estimated pose $\partA$.}
	\label{fig:hand-guiding-experiment}
\end{figure}

\begin{figure}[t]
	\centering
    \includegraphics[width=\columnwidth]{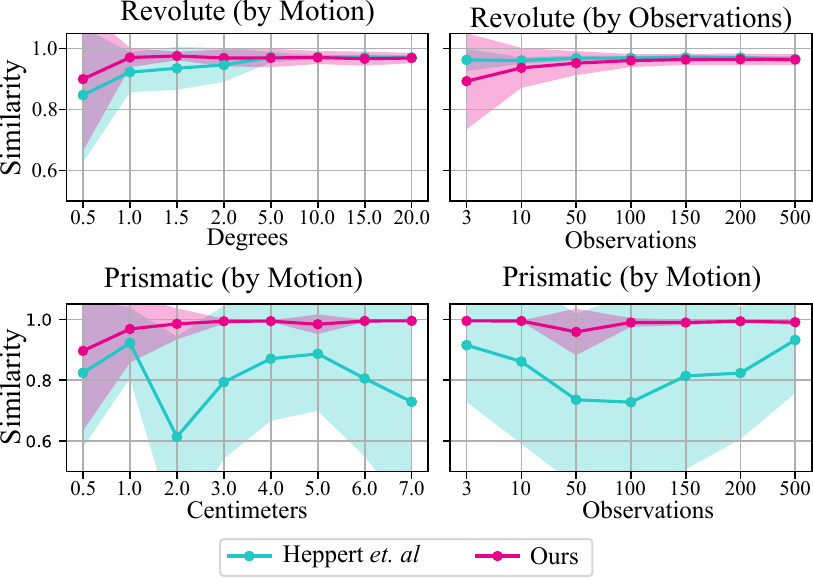}
	\caption{Tangent similarity for hand guiding experiments. The solid line shows average error, while the shaded region shows standard deviation.}
	\label{fig:hand-guiding-results}
\end{figure}

These experiments were conducted on the real robot hardware. We used the compliant KUKA LBR iiwa robot and physically attached the robot's end-effector to a box lid. We then hand-guided the robot motion in gravity compensation mode to open and close the box. For this experiment, we recorded both the robot joint positions, measured by the encoders, and the respective box lid poses, tracked with Vicon motion capture, as shown in Fig.~\ref{fig:hand-guiding-experiment}. 
Similar to \citet{Heppert2022}, we use the tangent similarity metric:
\begin{align}
    J(\linvel_{gt}, \linvel_{est}) = \frac{1}{\theta_{max} - \theta_{min}} \int_{\theta_{min}}^{\theta_{max}} \frac{\linvel_{gt}}{\norm{\linvel_{gt}}} \cdot \frac{\linvel_{est}}{\norm{\linvel_{est}}},
\end{align}
where $\linvel_{gt}$ is the local linear velocity of the grasp point measured from Vicon and $\linvel_{est}$ is the estimated local velocity from the articulation model. We can compute $\linvel_{est}$ from $\xi$ using the equation: $\linvel_{est} = \linvel + \rotvel \times \vec{c}$ where $\vec{c}$ is the contact point from kinematics. Since $\linvel_{gt}$ and $\linvel_{est}$ are normalized, they represent the direction of motion; therefore, their tangent similarity will be 1 when identical and 0 when perpendicular.

We recorded two hand guiding experiments, one for a revolute joint and one for a prismatic. First, we performed optimization over fixed increments, for example, optimizing over every \SI{1}{\degree} of rotation or \SI{1}{\centi\meter} of translation. Next, we tested using fixed numbers of measurements equally spaced over the entire configuration range, with full results shown in Fig.~\ref{fig:hand-guiding-results}. In the factor graph, we make no distinction between prismatic or revolute. When estimating prismatic joints, $\rotvel$ tends towards very small values. At the output, if $\norm{\rotvel} < 0.01$, we set $\rotvel = 0_{3\times1}$ and normalize $\linvel$.

\subsubsection{Results}
Our results demonstrate a high degree of accuracy, even with a small number of measurements. After only \SI{0.5}{\degree} of rotation, our estimator has an average tangent similarly of 0.90, after \SI{1.0}{\degree}, this improves to 0.97.
This enables online articulation estimation in cases where the neural network prediction is wrong because the robot part will only need to move the articulated part a small amount for the estimate to be updated.
Additionally, we show that for equally spaced measurements throughout the configuration range, as few as 3 measurements can be sufficient to accurately estimate the joint.
In comparison with Heppert~\etal, both methods have similar performance for revolute joints, while our method is better at distinguishing prismatic joints. We suspect this is because we check for prismatic articulations, whereas their method tends to confuse prismatic joints with very large revolute articulations.

\begin{figure*}[t]
	\centering
	\includegraphics[width=0.47\linewidth]{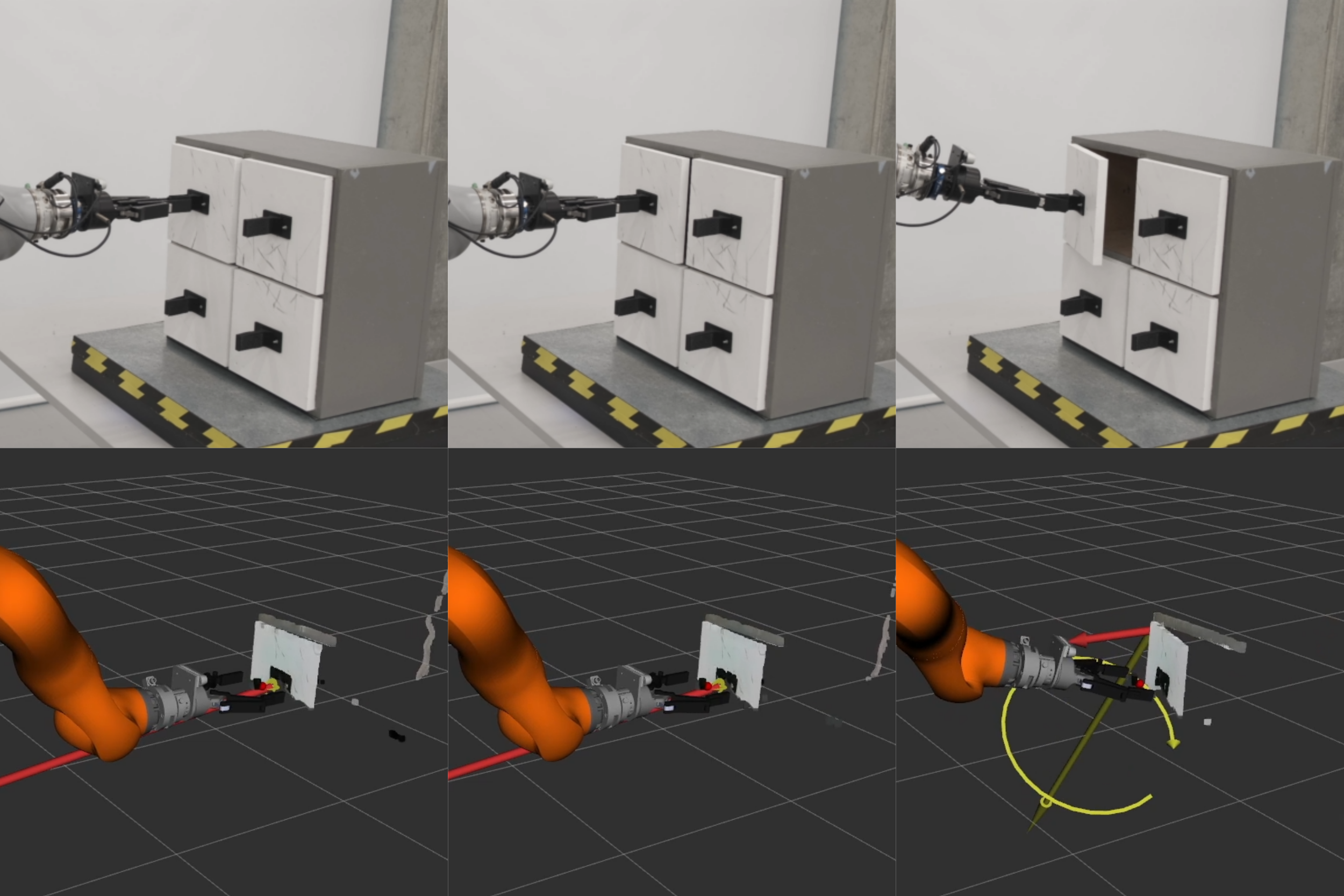}
    \includegraphics[width=0.47\linewidth]{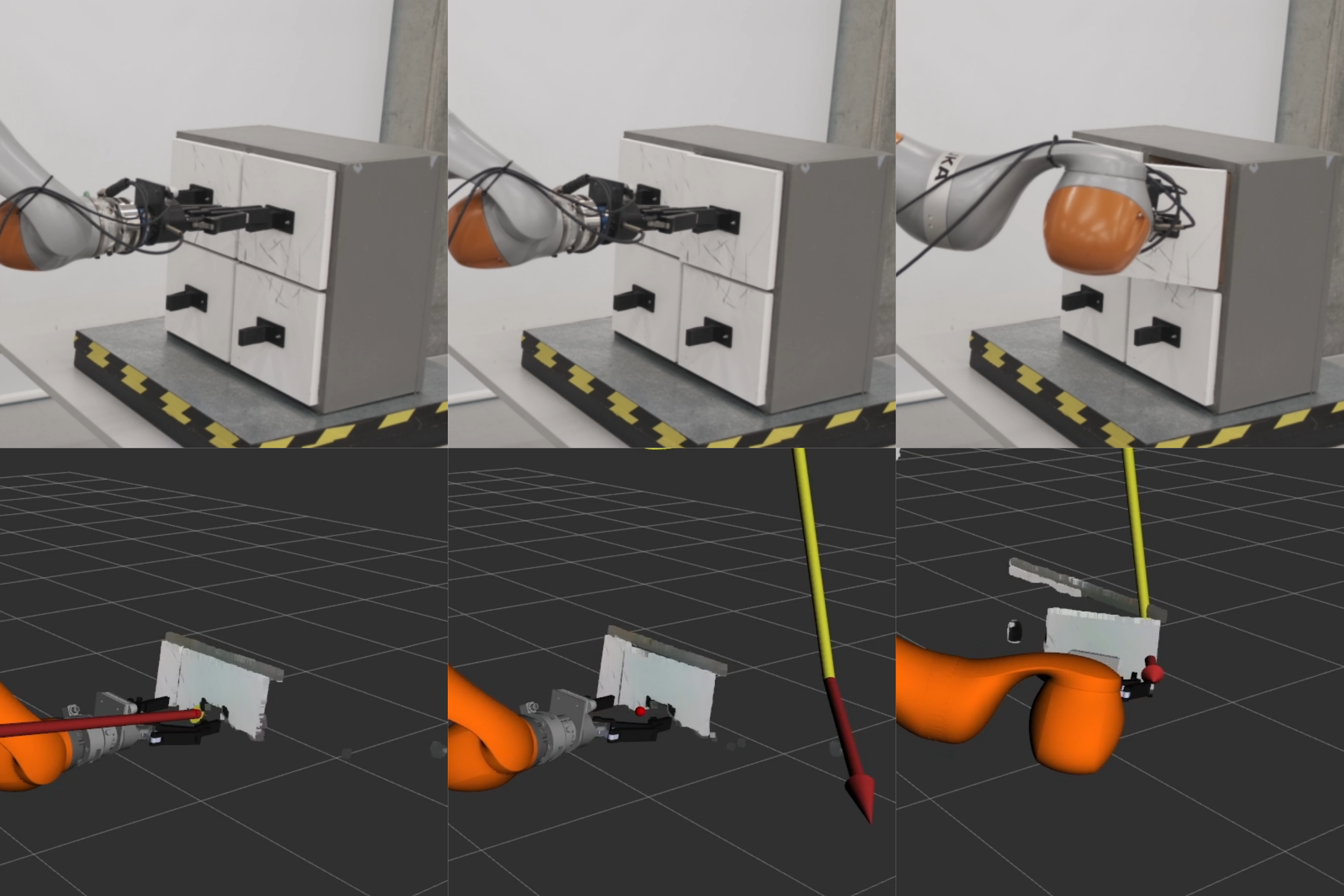}\\
    \vspace{2mm}
    \includegraphics[width=0.47\linewidth]{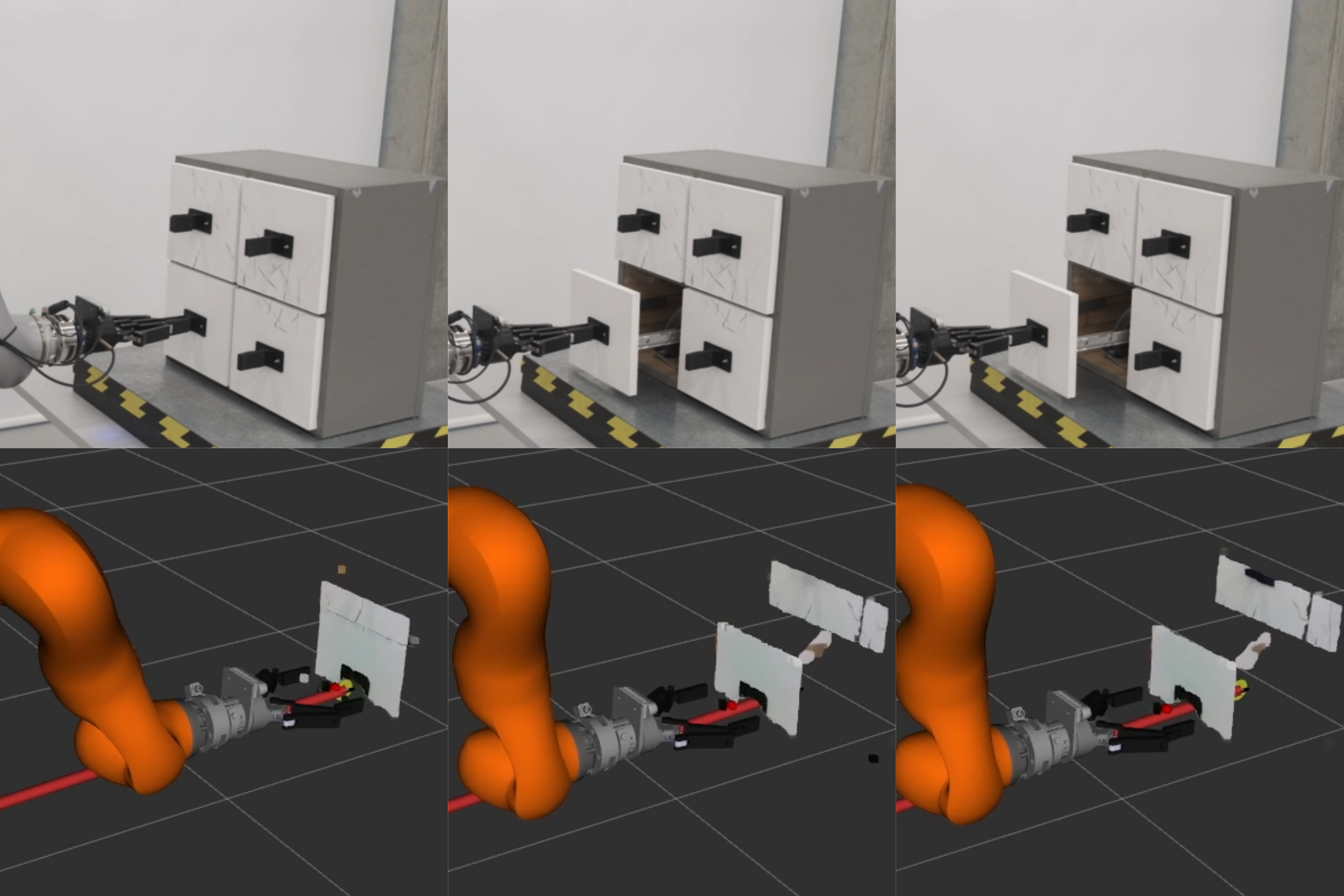}
    \includegraphics[width=0.47\linewidth]{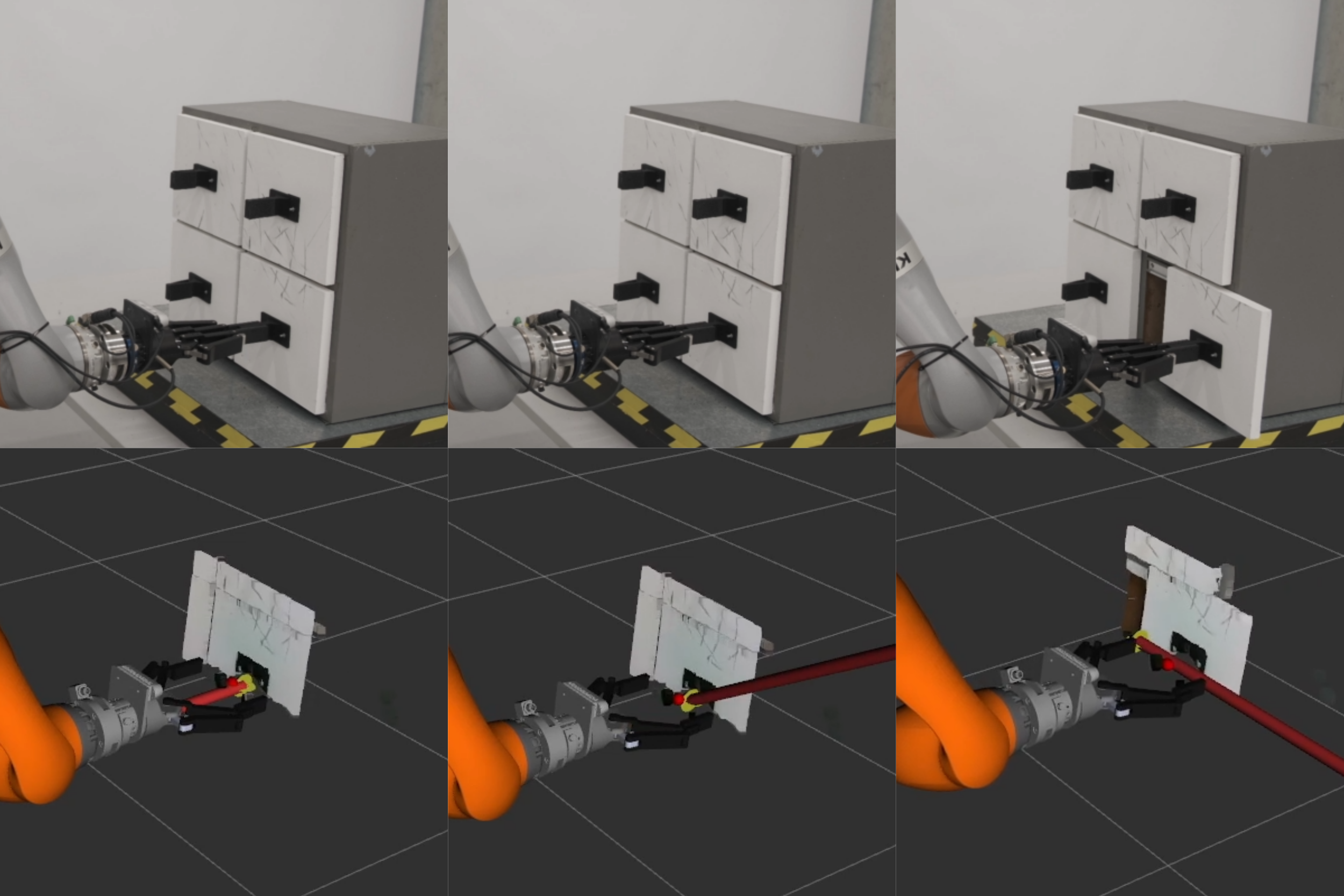}
	\caption{Robot experiments opening each of the four cabinet doors. \textbf{Top Left:} The network initially predicted a prismatic joint and as the arm pulled backwards on the door, the small amount of movement allowed the factor graph to solve for the correct revolute joint. \textbf{Top Right:} Similarly to the top left door, the network initially predicted prismatic but this estimated was updated online using the kinematic measurements. \textbf{Bottom Left:} the network correctly predicted prismatic joint and the robot easily opened the drawer. \textbf{Bottom Right:} the network predicted prismatic joint and the arm began to pull backwards. As the force threshold was passed, a force plane factor was added, which resulted in another prismatic joint prediction. This allowed the robot to move the door slightly, producing kinematic measurements that converged to the correct estimate.}
	\label{fig:robot-experiments}
\end{figure*}

\subsection{Shared-autonomy Robot Experiments}

For the shared-autonomy robot experiments, we used the same KUKA iiwa robot, and for sensing and grasping, we used an Intel RealSense D435 camera, an ATI Delta force/torque sensor, and a Robotiq 140 two-finger gripper. 
In these experiments, we tested the full pipeline as described in Sec.~\ref{sec:implementation} with the following experimental protocol: the human user views the robot's camera feed, which is looking at the same cabinet as in Fig.~\ref{fig:motivation-cabinet}, and clicks on the image where to gasp. The robot then moves to the grasp goal and closes the gripper. Next, the robot moves using the learned articulation prediction from $\theta = 0$ to a specified upper bound. 
If a force threshold is reached and the gripper has not moved, this triggers the addition of a force factor to the factor graph, which is then optimized to find the next MAP articulation. The robot arms again attempt to open the cabinet and are either successful or another force factor is added until the solution converges on a direction where the door begins to open.

As the estimation runs online, once the end-effector begins to open the door, even a small amount, kinematic sensing is added to the factor graph, and the model is updated. This is fed back to the controller in a closed loop. Eventually, the motion of the arm allows more of the door to open, which leads to more kinematic measurements, and the estimate converges to the correct estimate of the joint, and the controller continues to open and close the door. We performed a new optimization after every 20 new kinematic measurements and used a distance limit of $d=\SI{2}{\milli\meter}$ or $d=\SI{0.5}{\degree}$ to trigger adding a new kinematic measurement to the factor graph. 

\subsubsection{Results}
\begin{figure*}[t]
	\centering
	\includegraphics[width=\linewidth]{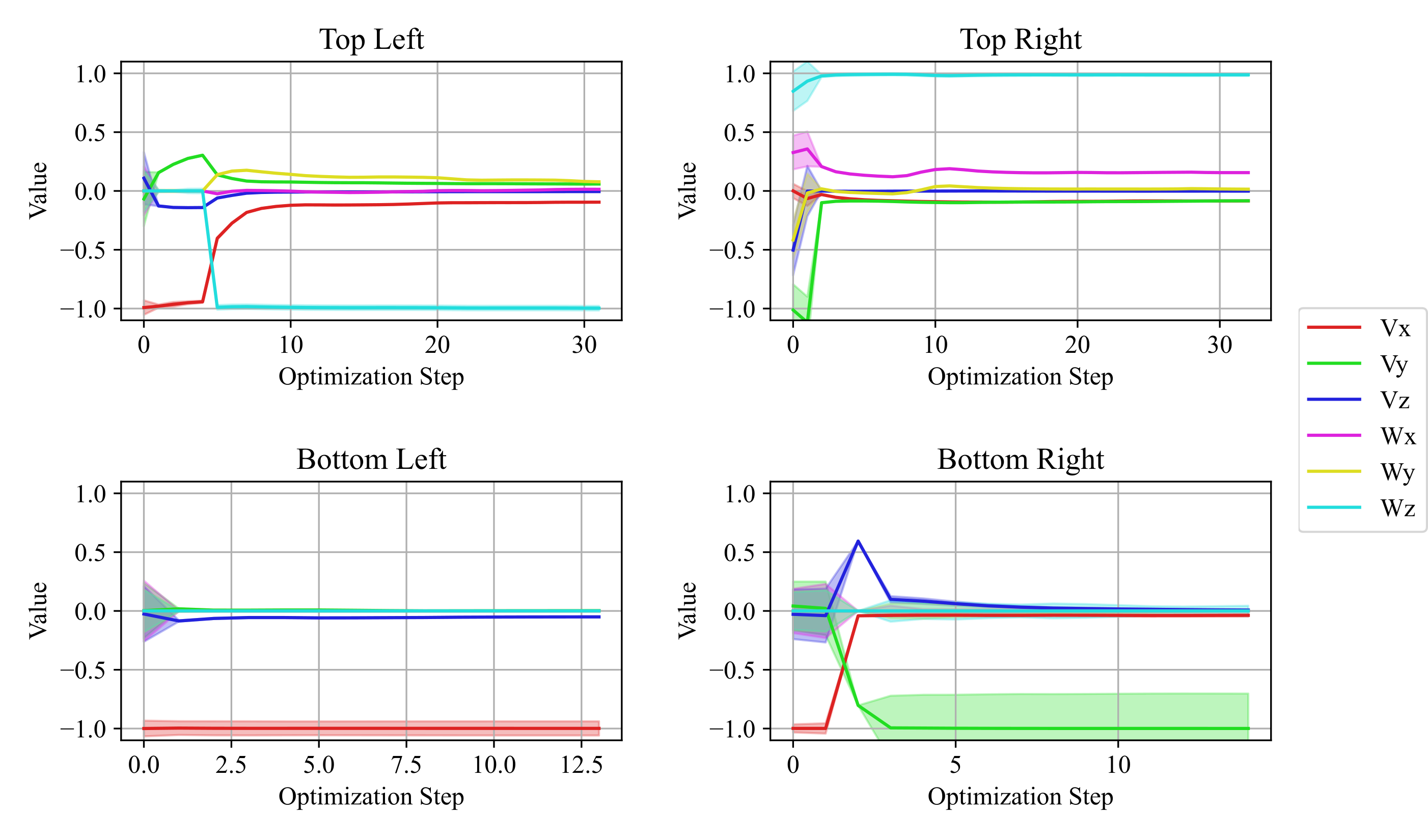}
	\caption{Plots of the estimated $\xi$ parameters during the experiments in Fig.~\ref{fig:robot-experiments} experiment and the associated $3\sigma$ computed from marginal covariances.}
	\label{fig:robot-estimation-results}
\end{figure*}

Four of the online estimation experiments are shown in Fig.~\ref{fig:robot-experiments}. As similarly shown in previous work~\cite{Buchanan2024}, without visual cues for articulation, affordance-based neural networks tend to predict prismatic joints. In this work, because of our articulation prediction factor and the inclusion of the force plane factor, the robot was able to update the estimate even in the bottom right sliding door case. We repeated the full pipeline experiment 20 times on different doors and successfully opened the doors 15 times. Fig.~\ref{fig:robot-estimation-results} shows the estimated $\xi$ parameters during online experiments. Marginal covariances are computed for each optimization, and the $3\sigma$ range is depicted in Fig.~\ref{fig:robot-estimation-results} as a shaded area around the estimated mean value.

\section{Discussion}
\label{sec:discussion}
Our method involves several advances that enable robots to open visually ambiguous objects of unknown articulation. First, by changing the neural network to provide a prediction of uncertainty and by changing the articulation factor, we enabled the initial prediction of articulation to include a learned uncertainty distribution rather than a hard-coded one as before. Fig.~\ref{fig:covariance-example} shows an example network output of a sliding door with covariance largest along the $x$ direction, followed by $y$. 

The addition of force sensing into the factor graph allows for the uncertainty to be used to update the estimate to the next most likely articulation. As an example in Fig.~\ref{fig:covariance-example}, the dominant prediction from the network is of a revolute door, however, it is clear from the uncertainty distribution that the network is less confident about lateral motion. When a force plane factor is added to the factor graph, the estimate will collapse along the $y$ direction, correctly updating the estimate as prismatic. This is the process that allowed the robot to open the bottom right sliding door in Fig.~\ref{fig:robot-experiments}.
Of the 20 full system trials attempted, 4 failed due to the slipping of the gripper. Because we rely on an assumption of rigid contact with only a small amount of slipping, we cannot differentiate between significant slipping and intentional movement opening a door. In future work, this could be detected using sensors on the fingertips of the gripper.

In our experiments, we found that the compliance in the robot arm could significantly affect the success rate. For example, if the robot arm is too compliant in a specific direction, it may not be able to overcome the friction in the joint to open the door. On the other hand, if the arm is too stiff, it could break the door. This presents an opportunity for future investigation by adapting the stiffness parameters online based on the estimate of the articulation. Model-based or learning-based approaches may equally be applied to this problem. Learning-based methods also present a path for learning priors for safe interaction forces. As it stands, the maximal force exerted by our system is a hyperparameter that has to be adjusted for the setup. While cabinet-sized objects all require very similar interaction forces, we would of course like a system that can also operate heavy doors and small jewelry boxes without the need for human intervention. We see a path for learning such priors on the basis of stable language-aligned visual features~\cite{radford2021learning}, either incrementally or from human demonstrations~\cite{li2022estimating}.

Finally, we use force sensing only at the beginning of the interaction for estimating the articulation; however, in future work, we intend to explore learning-based methods for estimating articulation from force sensing, similar to learning pose estimators~\cite{ferrandis2024}.
Despite the Sim2Real gap for contact physics, we believe that leveraging training in simulation~\cite{aoyama2024} can significantly improve success rates for real-world training of robot interactions with articulated objects.

\section{Conclusion}
\label{sec:conclusion}
In this work, we present a novel method for online estimation and opening of unknown articulated objects. Our method can enable a robot to open a wide variety of articulated objects including both common household items and objects whose articulation is not visually apparent. Our method fuses visual, force and kinematic sensing from both learned predictions from a neural network, and physics-based modeling of articulations using screw theory. The back-bone estimation framework is based on factor graphs which is integrated with a shared autonomy framework in which a user simply clicks where to open, and the robot opens a door.

This work significant expand on our previous work with several major advances. We modified the neural network to provide a prediction of uncertainty, and we introduced a new articulation factor to facilitate the incorporation of this uncertainty into the factor graph. We also added an entirely new sensing modality in force sensing. The combination of these changes made our system much more capable of opening different articulations, including where the articulation is not visually apparent.
We implemented our method on a real robot for interaction with a visually ambiguous articulated object and achieved a high rate of success for interaction. While more work can be done on integrating force sensing into the framework, this article shows the major benefits of fusing proprioceptive sensing with learned vision priors for object manipulation.

\bibliographystyle{spbasic}
\bibliography{ref}

\end{document}